\PassOptionsToPackage{prologue,dvipsnames}{xcolor}
\documentclass[10pt,twocolumn,letterpaper]{article}

\usepackage[pagenumbers]{cvpr} 

\definecolor{cvprblue}{rgb}{0.21,0.49,0.74}
\usepackage[pagebackref,breaklinks,colorlinks,allcolors=cvprblue]{hyperref}

\usepackage{hyperref}
\usepackage{url}
\usepackage{booktabs}       
\usepackage{amsfonts}       
\usepackage{nicefrac}       
\usepackage{microtype}      
\usepackage{bm}
\usepackage{enumitem}
\usepackage{amsmath,amsthm,amssymb}
\usepackage{mathtools}
\usepackage{multirow}
\usepackage{outlines}
\usepackage{enumitem}
\usepackage{caption}
\usepackage{subcaption}
\usepackage{algorithm}
\usepackage{listings}
\usepackage{algpseudocode}
\usepackage{arydshln}
\usepackage{wrapfig,booktabs}
\usepackage{lipsum}

\hypersetup{colorlinks,citecolor={blue}}

\newcommand{\Hquad}{\hspace{0.25em}} 

\definecolor{Gray}{gray}{0.9}

\def\rvx{{\mathbf{x}}}
\def\rvy{{\mathbf{y}}}

\title{HFI: A unified framework for training-free detection and implicit watermarking of latent diffusion model generated images}

\author{Sungik Choi\\
LG AI Research\\
{\tt\small sungik.choi@lgresearch.ai}
\and 
Hankook Lee\\
Sungkyunkwan University\\
\and
Jaehoon Lee\\
LG AI Research\\
\and 
Robin Kim\\
University of Massachusetts at Amherst\\
\and 
Stanley Jungkyu Choi\\
LG AI Research\\
\and 
Moontae Lee\\
LG AI Research\\
}

\usepackage{xspace}
\newcommand{\methodname}{$\text{HFI}$\xspace}

\begin{document}

\maketitle
\begin{abstract}

Dramatic advances in the quality of the latent diffusion models (LDMs) also led to the malicious use of AI-generated images. While current AI-generated image detection methods assume the availability of real/AI-generated images for training, this is practically limited given the vast expressibility of LDMs. This motivates the training-free detection setup where no related data are available in advance. The existing LDM-generated image detection method assumes that images generated by LDM are easier to reconstruct using an autoencoder than real images. However, we observe that this reconstruction distance is overfitted to background information, leading the current method to underperform in detecting images with simple backgrounds. To address this, we propose a novel method called HFI. Specifically, by viewing the autoencoder of LDM as a downsampling-upsampling kernel, HFI measures the extent of aliasing, a distortion of high-frequency information that appears in the reconstructed image. HFI is training-free, efficient, and consistently outperforms other training-free methods in detecting challenging images generated by various generative models. We also show that HFI can successfully detect the images generated from the specified LDM as a means of implicit watermarking. HFI outperforms the best baseline method while achieving magnitudes of speedup.

\end{abstract}
\section{Introduction}
\label{sec:intro}

With the rapid advancement of generative AI, we are now able to generate photorealistic images in a desired context within seconds. The crucial factor in this achievement is the emergence of foundational Latent Diffusion Models (LDMs) \citep{22Midjourney, Rombach22SD}. LDMs represent the integration of recent breakthroughs of vision models, including the powerful generation quality of diffusion models \citep{Song19Scorebased,Song21SDE}, the joint text-image representation learned by vision-language pre-trained models \citep{Radford21CLIP}, and the improved inference speed through compression in the latent space \citep{Kingma14VAE}. However, with the progression of such models, we should also consider the potential negative impacts such generated images may have on society. For example, LDMs can produce fake images that could cause societal confusion \citep{retuers24southkorea} or infringe upon intellectual property rights \citep{reuters24lawsuit}. Moreover, applying AI-generated images on training generative models may deteriorate model performance \citep{Alemohammad24Madcow}. Therefore, AI-generated image detection or deepfake detection methods, which aim to distinguish AI-generated images from real images, are gaining significant attention.

\begin{figure*}[t]
\centering
\includegraphics[width=0.96\textwidth]{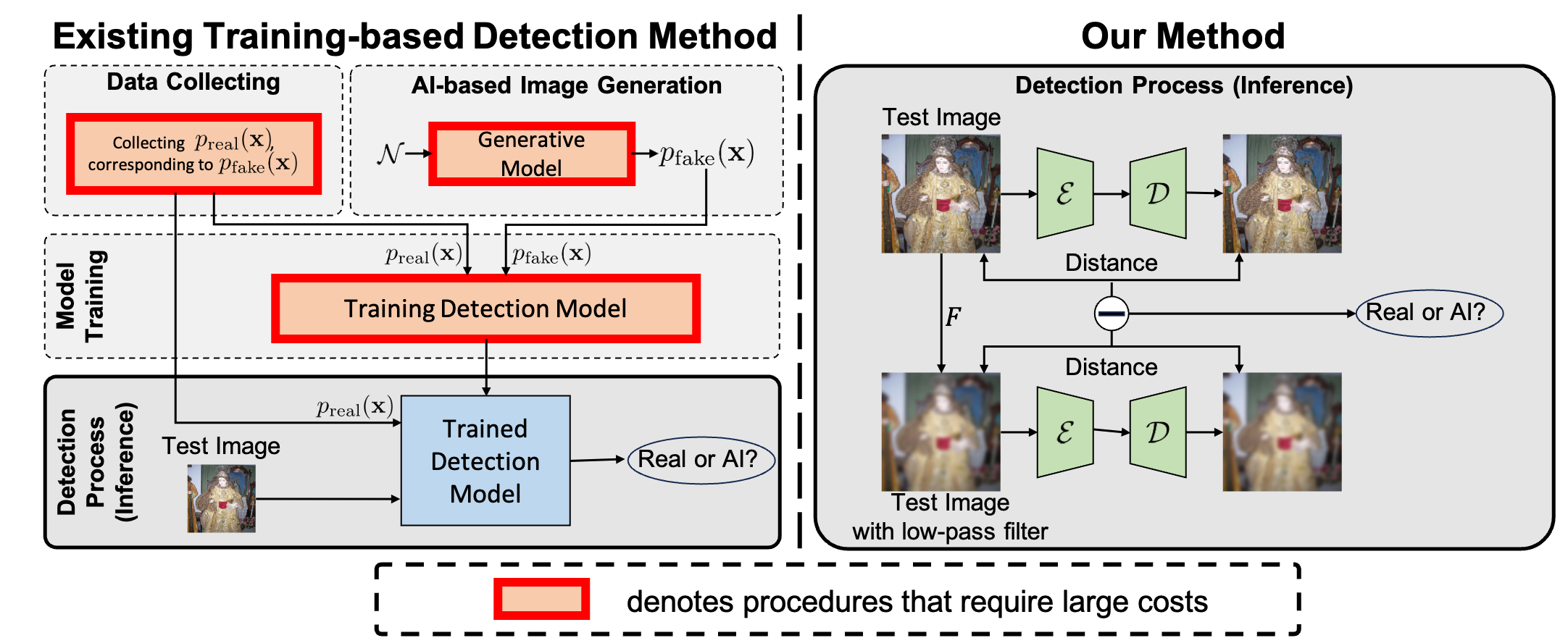}
\caption{\textbf{Problem setup of \methodname.} \textbf{(Left)} Setup of training-based AI-generated image detection methods. Such methods train and test on the same real data distribution. Furthermore, the framework can be costly when detecting images produced by large-scale text-to-image generative models.  \textbf{(Right)} Pipeline of our proposed \methodname. \methodname operates only on the test time and can be computed efficiently via the autoencoder of the LDM.}
\label{fig:intro}
\end{figure*}

However, most AI-generated image detection methods adhere to traditional settings that are disconnected from real-world challenges. As shown in Figure \ref{fig:intro}, these methods follow a training-based setup. Namely, they are designed to train on data sampled from a given real image distribution with similar AI-generated data, and they are tested on the data sampled from the same real image distribution against other AI-generated data. While such a setup was effective when generative models were typically designed to fit a single specific dataset, it faces practical limitations when we aim to detect LDM-generated data in general, which are trained on billions of images \citep{Schumann22LAION}. Moreover, such LDMs can generate hallucinated images that may not have been encountered in reality \citep{guardian2024googleai}, which induces extra effort to acquire matching real images.

We explore a training-free AI-generated image detection framework that assumes no access to training images as an alternative to traditional setups. A few methods in this domain \citep{ricker24aeroblade, He24rigid} aim to design a universal score function that distinguishes AI-generated images from real ones by leveraging representations from large-scale pre-trained models. For example, \citet{ricker24aeroblade} and \citet{He24rigid} leverage the representation of the autoencoder of LDM and Dinov2 \citep{oquab24dinov2}, respectively. While these methods are directly deployable in real-world scenarios without further training, none of the methods have shown their efficacy on challenging benchmarks (\eg, GenImage \citep{Zhu23genimage}) that contain LDM-generated images, or images generated by other text-to-image generative models (\eg, GLIDE \citep{Nichol22GLIDE}).\\

\noindent
\textbf{Contributions.} In this paper, we first challenge the pitfalls of the best baseline method \citep{ricker24aeroblade} that utilizes reconstruction distance when applied to broader images. We show that even when the right autoencoder is used, the reconstruction can be easier in the real data with simpler backgrounds (Figure \ref{fig:toy:ab}). Based on our observations that autoencoders of the LDMs fail to reconstruct the high-frequency components of real-image, we propose our method, High-frequency influence (\methodname). \methodname reduces the impact of background information by taking account of high-frequency components. As shown in Fig \ref{fig:intro}, our method can be computed efficiently during test time. It is worth noting that our method can also incorporate various design choices on the distance and high-frequency filters.


We demonstrate the efficacy of \methodname through extensive experiments in challenging AI-generated image detection benchmarks in various domains (\eg,  natural images \citep{Zhu23genimage}, face \citep{Chen24diffusionface}). \methodname outperforms existing training-free methods consistently. Moreover, HFI is competitive with the state-of-the-art training-based method, DRCT \citep{Chen24DRCT}.

We also apply \methodname on tracing model-generated images given a specific LDM model, where a reliable uncertainty metric is required to distinguish model-generated images from the other effectively. While such a metric can effectively trace the image ownership without explicit watermarking, the current state-of-the-art method \citep{Wang24LatentTracer} is based on input optimization, which is computationally inefficient. Our \methodname outperforms the baseline method while achieving magnitudes of speedup.

In brief, our contributions are summarized as follows.
\begin{itemize}[leftmargin=15pt,topsep=0pt]
\setlength\itemsep{0em}
\item We propose \methodname, a novel score function that distinguishes AI-generated images from real images without any training (Section \ref{sec:method}).

\item \methodname outperforms existing baselines on challenging benchmarks (Section \ref{ssec:mainresults}). We conduct extensive ablation studies for potential improvement of \methodname (Section \ref{ssec:ablationstudy}).

\item We also apply \methodname into the distinct setup of detecting LDM-generated images from the specified LDM (Section \ref{sec:ldmtrace}). \methodname improves over the  baseline \citep{Wang24LatentTracer} with significant speedup.
\end{itemize}

\section{Preliminaries}
\label{sec:background}

\subsection{Problem Setup}
\label{ssec:problemsetup}

Given the real data distribution $p_{\text{real}}(\rvx)$, the manipulator can utilize the generative model $G_{i}$ to produce deepfake distribution $p_{\text{fake},G_{i}}(\mathbf{x})$ that is similar to $p_{\text{real}}(\mathbf{x})$. The goal of the AI-generated image detection is to design a score function $U(\mathbf{x})$ that determines whether $\mathbf{x}$ is from $p_{\text{real}}(\mathbf{x})$ (\ie, $U(\mathbf{x}) > \tau$) or not (\ie, $U(\mathbf{x}) \le \tau$). We denote ``$\mathtt{A}$ \textit{vs} $\mathtt{B}$'' as an AI-generated image detection task where $\mathtt{A}$ is the real dataset and $\mathtt{B}$ is the applied generative model to mimic dataset $\mathtt{A}$. For clarity, we note that the term $"\mathtt{real \Hquad image}"$ in this paper does not include samples in the training dataset of $\mathtt{B}$.

This paper studies training-free AI-generated image detection where universal metric design is required without prior knowledge. Namely, we do not have access to $p_{\text{real}}(\mathbf{x})$ or $p_{\text{fake},G_{i}}(\mathbf{x})$. Given the diverse expressibility of concurrent text-to-image generation models (\eg, Stable Diffusion (SD) \citep{Rombach22SD}), our setup matches the goal of real-world AI-generated image detection.

\begin{figure*}[ht]
\centering

\begin{subfigure}[b]{0.24\textwidth}
\includegraphics[width=0.92\textwidth]{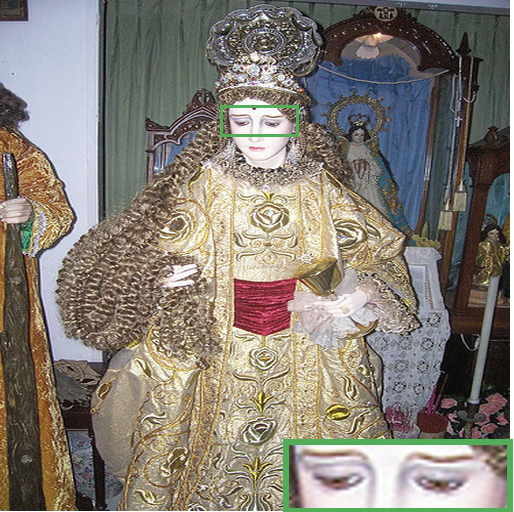}
\caption{Original image}\label{fig:toy:org}
\end{subfigure}
\hfill
\begin{subfigure}[b]{0.24\textwidth}
\includegraphics[width=0.92\textwidth]{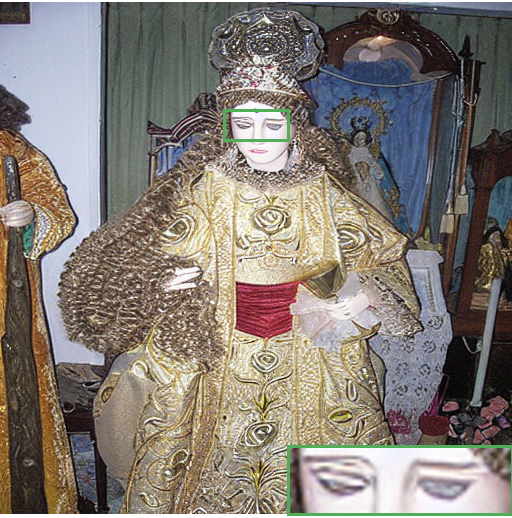}
\caption{Reconstructed image}\label{fig:toy:recon}
\end{subfigure}
\hfill
\begin{subfigure}[b]{0.24\textwidth}
\includegraphics[width=0.92\textwidth]{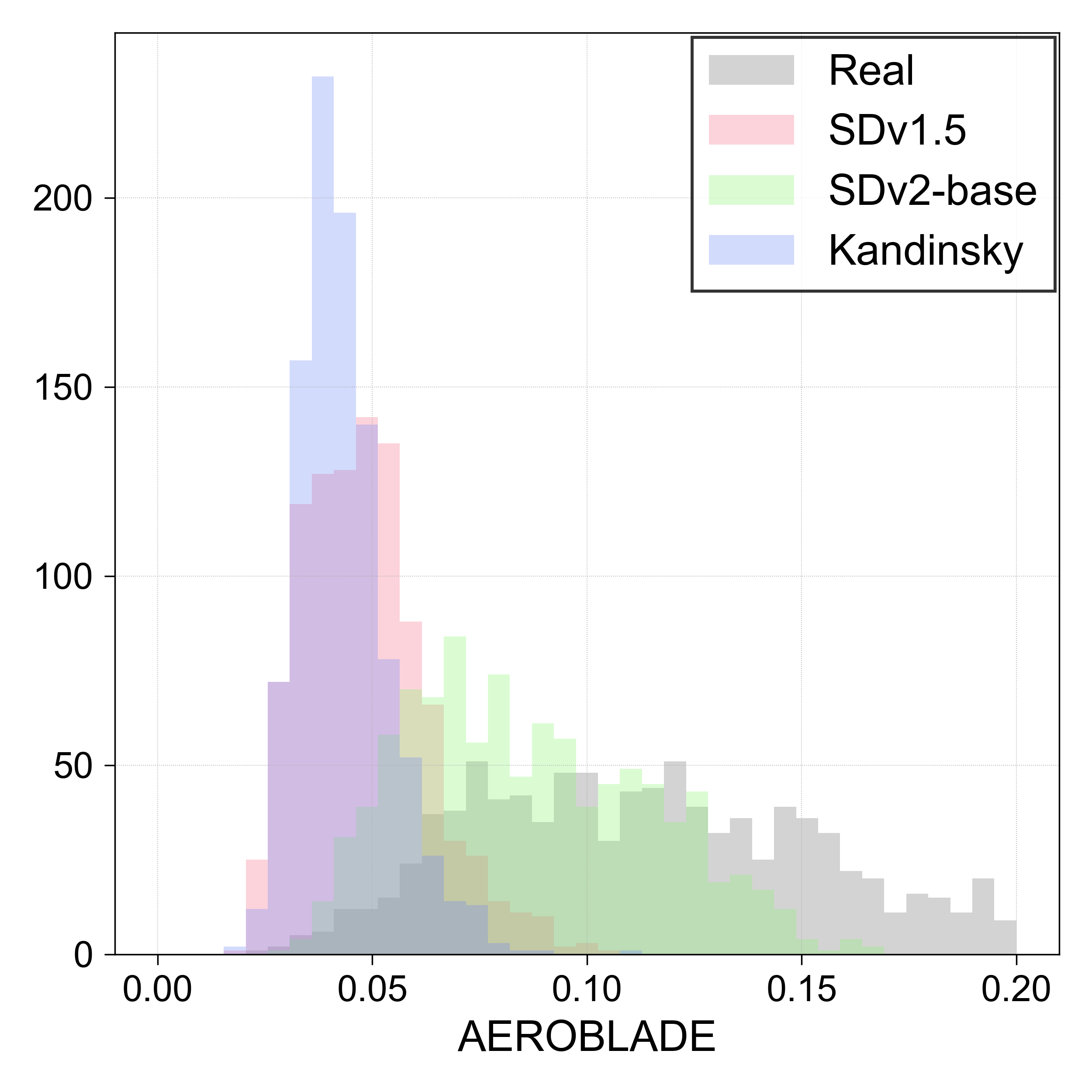}
\caption{Histogram of AEROBLADE}\label{fig:toy:ab}
\end{subfigure}
\begin{subfigure}[b]{0.24\textwidth}
\includegraphics[width=0.92\textwidth]{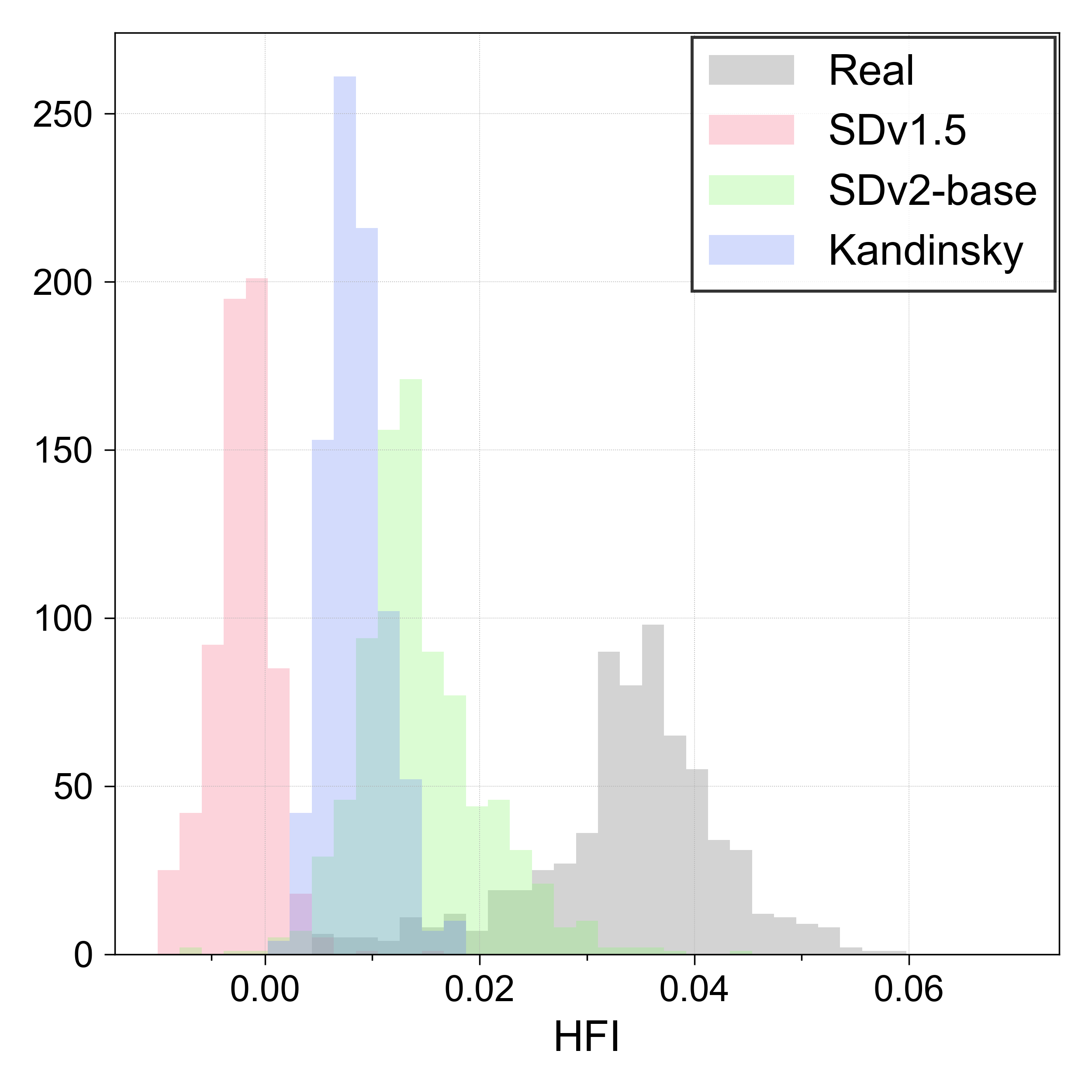}
\caption{Histogram of \methodname}\label{fig:toy:hfi}
\end{subfigure}
\caption{\textbf{Motivation of \methodname.} \textbf{(a)} Sampled data from the ImageNet \citep{Deng09ImageNet} dataset. \textbf{(b)} Reconstruction through the autoencoder of the Stable Diffusion \citep{Rombach22SD} v1.1 model. We can observe obvious distortions in the high-frequency details. \textbf{(c)} Histogram of AEROBLADE \citep{ricker24aeroblade} experimented in toy dataset. \textbf{(d)} Histogram of \methodname experimented in toy dataset.}
\label{fig:motivation}
\end{figure*}
\begin{figure}[ht]
\centering
\includegraphics[width=0.45\textwidth]{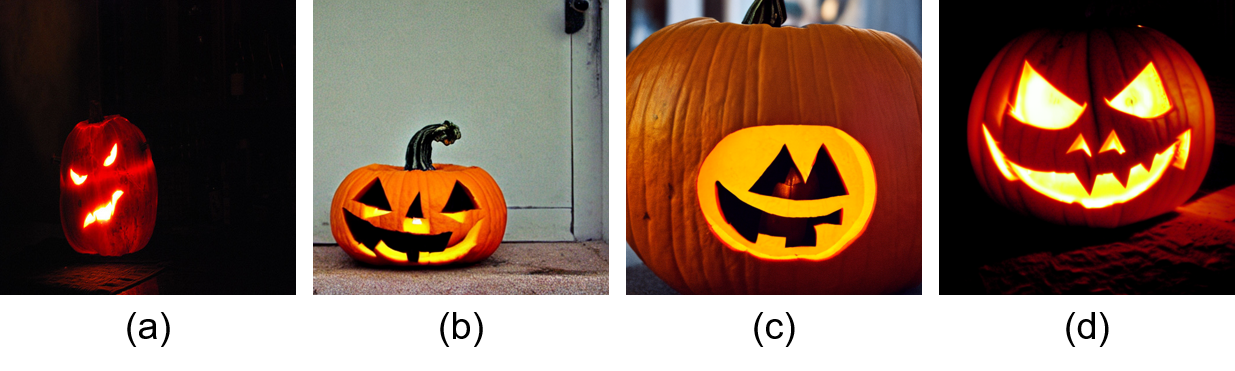}
\caption{\textbf{Samples of toy dataset.} \textbf{(a)} Real ImageNet \citep{Deng09ImageNet} data. \textbf{(b),(c),(d)} AI-generated data. SDv1.5 \textbf{(b)}, SDv2-base \textbf{(c)} \citep{Rombach22SD}, and Kandinsky \textbf{(d)} \citep{Razzhigaev23Kandinsky} are applied for generation, respectively.}
\label{fig:lantern}

\end{figure}

\subsection{Latent diffusion models}
\label{ssec:ldmbasic}

 LDMs \citep{Rombach22SD} efficiently generate high-dimensional images by modeling the diffusion process on the latent space $\mathcal{Z} \subset \mathbb{R}^{C^{\prime} \times H^{\prime} \times W^{\prime}}$ instead of the data space $\mathcal{X} \subset \mathbb{R}^{C \times H \times W}$ with $C=3$. Hence,  LDMs first pre-train the encoder $\mathcal{E}: \mathcal{X} \rightarrow \mathcal{Z}$ and the decoder $\mathcal{D}: \mathcal{Z} \rightarrow \mathcal{X}$. This paper will refer to LDM as a comprehensive framework for autoencoder-based text-to-image generative models.
 
 While the autoencoder of the LDM is defined as VAEs \citep{Kingma14VAE}, weight on the KL-regularization loss during training is negligible, resulting in negligible variance in the latent space \citep{Rombach22SD}. Hence, we view the encoder and the decoder as the deterministic mapping between $\mathcal{X}$ and $\mathcal{Z}$ throughout the paper. We set $\text{AE}(\rvx) \coloneq \mathcal{D} \circ \mathcal{E}(\rvx)$ for the rest of the paper. 

 \subsection{Attribution of LDM-generated images.}
\label{ssec:ltracebasic}

  A recent line of works \citep{Wang23RONAN, Wang24LatentTracer} aims to detect data generated from the given LDM model $\mathcal{M}$ as a means of alternative to explicit watermarking \citep{Fernandez23Stablesignature, Wen23Treering}. In specific, for a given LDM model $\mathcal{M}_{1}$, existing methods design an uncertainty score function that distinguishes the belonging data generated from $\mathcal{M}_{1}$ and the non-belonging data generated from the other generative model $\mathcal{M}_{2}$. We denote such task as ''$\mathcal{M}_{1}$ $\mathtt{vs}$ $\mathcal{M}_{2}$'' model attribution task.

\section{Method}
\label{sec:method}

\subsection{Motivation}
\label{ssec:motivation}

In this subsection, we challenge the pitfalls of the AE-based baseline method, AEROBLADE \citep{ricker24aeroblade}, when applied to challenging real-world cases. Specifically, we show its vulnerability to local style variation, where the real data has simpler backgrounds. For examination, we sample 1000 data of the class "$\mathtt{Jack-o'-lantern}"$ from the ImageNet \citep{Deng09ImageNet} dataset and generate the image via the text prompt "$\mathtt{A\Hquad photo\Hquad of\Hquad a\Hquad Jack-o'-lantern}$". We use the SDv1.5, SDv2-base \citep{Rombach22SD}, and Kandinsky \citep{Razzhigaev23Kandinsky} models for the generation.  We present the representative sampled image in Figure \ref{fig:lantern}. The ImageNet and Kandinsky-generated images show monotonous backgrounds that are easy to reconstruct.

We now verify the AEROBLADE on this toy dataset. Specifically, given the input data $\rvx$, AEROBLADE measure the LPIPS \citep{zhang18LPIPS} distance between $\rvx$ and $\text{AE}(\rvx)$. For the autoencoder, we use SDv1.5 for examination. We present the results in Figure \ref{fig:toy:ab}. Simple reconstruction distance struggles to detect between SDv2-base-generated images against ImageNet images and SDv1.5-generated images against Kandinsky-generated images. Each case shows the baseline's pitfalls on the zero-shot image detection and attribution of LDM-generated images, respectively. Specifically, AEROBLADE outputs smaller distances to images with simpler backgrounds.

\subsection{Methodology}

We discuss the score function that reduces the bias to the background information. Our motivation is grounded in the phenomenon of aliasing \citep{Gonzalez06DIP}, which occurs when an original signal contains high-frequency components exceeding the subsampling rate, leading to distortions in the downsampled signal and subsequently affecting its upsampled reconstruction. Since the autoencoder in the LDMs can be viewed as implicit downsampling-upsampling kernels, such aliasing can occur on the reconstructed data when the novel image is given as the input. In this work, we propose that the extent of aliasing present in reconstructed images can serve as a reliable metric for distinguishing between real and AI-generated images. 



We show our motivation with the reconstruction of the real image. Figure \ref{fig:toy:org} and \ref{fig:toy:recon} show the sample of ImageNet \citep{Deng09ImageNet} and its reconstruction through the SDv1.1 autoencoder, respectively. The encoder $\mathcal{E}$ fails to compress high-frequency components of the real image and causes deviation. For example, we can observe obvious distortions in the high-frequency details in the reconstructed images (\eg, eye, ring, pattern in cloths). 


Motivated by the observations, we propose to measure the influence of the input high-frequency components on the discrepancy between the input data and its reconstruction as a detection score function. We propose our \methodname as follows




\begin{equation}\label{eq:sungwoo_HFI}
\text{HFI}_{d,\mathcal{F}, \text{AE}, \nu}(\rvx) \coloneq  \left\langle \frac{\partial d(\rvx,\text{AE}(\nu, \rvx))}{\partial \rvx} \; , \; \rvx-\mathcal{F}(\rvx)  \right\rangle  
\end{equation}
where $d:\mathcal{X} \times \mathcal{X} \rightarrow \mathbb{R}_{0}^{+}$ is the reconstruction distance function and $\mathcal{F}: \mathcal{X} \rightarrow \mathcal{X}$ are low-pass filters, where the autoencoders $(\ie, \text{AE}(\rvx))$ are trained on original dataset $\nu$:
\begin{equation}
    \text{AE}(\nu, \cdot) = \arg\min_{\mathcal{D}, \mathcal{E}} \mathbb{E}_{\rvy \sim \nu}\left[\big |\big|\mathcal{D} \circ \mathcal{E}(\mathbf{\rvy}) - \rvy \big |\big|^2\right] 
\end{equation}

\begin{table*}[t]
\centering
\caption{Mean AI-generated image detection performance (AUROC/AUPR) and average rank of \methodname and AEROBLADE \citep{ricker24aeroblade} in the GenImage  \citep{Zhu23genimage} dataset under the cross-autoencoder setup. \textbf{Bold} denotes the best method.}
\label{table:crossae_genimage_main}
\resizebox{\textwidth}{!}{
\begin{tabular}{cccccccccc}
\toprule
& \multicolumn{2}{c}{AE: SDv1.4} & \multicolumn{2}{c}{AE: SDv2-base} & \multicolumn{2}{c}{AE: Kandinsky} & \multicolumn{2}{c}{AE: MiniSD}\\ \cmidrule(lr){2-3}\cmidrule(lr){4-5}\cmidrule(lr){6-7}\cmidrule{8-9}  
Method  & Mean & Avg Rank & Mean & Avg Rank & Mean & Avg Rank & Mean & Avg Rank \\ 
\midrule
$\text{AEROBLADE}_{\text{LPIPS}}$ & 0.897/0.891 & 3.63/3.75 & 0.846/0.848 & 3.63/3.75 & 0.782/0.783 & 3.63/3.63 & 0.855/0.839 & 3.75/3.63\\
$\text{AEROBLADE}_{\text{LPIPS}_{2}}$ & 0.928/0.927 & 2.88/2.75 & 0.873/0.882 & 3.38/3.25 & 0.810/0.819 & 3.38/3.38 & 0.882/0.867 &  2.88/2.88\\
$\methodname_{\text{LPIPS}}$ (ours) & 0.946/0.937 & \textbf{1.75}/2.00 & 0.916/0.917 & 1.75/1.75 & \textbf{0.895}/\textbf{0.900} & \textbf{1.38}/\textbf{1.50} & 0.927/0.925& 1.88/2.00\\
$\methodname_{\text{LPIPS}_{2}}$ (ours) & \textbf{0.959}/\textbf{0.965} & \textbf{1.75}/\textbf{1.50} & \textbf{0.923}/\textbf{0.936} & \textbf{1.25}/\textbf{1.25} & 0.880/0.897 & 1.63/\textbf{1.50} & \textbf{0.930}/\textbf{0.927} & \textbf{1.50}/\textbf{1.50}\\

\bottomrule
\end{tabular}}
\end{table*}

\begin{table*}[t!]
\centering
\caption{Mean AI-generated image detection performance (AUROC/AUPR) and average rank (lower is better) of \methodname and AEROBLADE \citep{ricker24aeroblade} in the DiffusionFace \citep{Chen24diffusionface} dataset under the cross-autoencoder setup. \textbf{Bold} denotes the best method.}
\label{table:crossae_diffusionface_main}
\resizebox{\textwidth}{!}{
\begin{tabular}{cccccccccc}
\toprule
 & \multicolumn{2}{c}{AE: SDv1.4} & \multicolumn{2}{c}{AE: SDv2-base} & \multicolumn{2}{c}{AE: Kandinsky} & \multicolumn{2}{c}{AE: MiniSD}\\ \cmidrule(lr){2-3}\cmidrule(lr){4-5}\cmidrule(lr){6-7}\cmidrule{8-9}  
Method  & Mean & Avg Rank & Mean & Avg Rank & Mean & Avg Rank & Mean & Avg Rank \\ 
\midrule
$\text{AEROBLADE}_{\text{LPIPS}}$ & 0.750/0.730 & 2.75/2.75 & 0.709/0.688 &  3.00/2.75 & 0.651/0.631 & 2.75/2.75 & 0.601/0.583 & 3.25/3.25\\
$\text{AEROBLADE}_{\text{LPIPS}_{2}}$ & 0.729/0.713 & 3.50/3.25 & 0.705/0.684 & 3.75/3.50 & 0.648/0.627 & 3.75/3.50 & 0.614/0.592 & 2.75/2.75\\
$\methodname_{\text{LPIPS}}$ (ours) & \textbf{0.772}/\textbf{0.770} & \textbf{1.75}/\textbf{1.75} & 0.732/\textbf{0.729} & \textbf{1.50}/\textbf{1.50} & \textbf{0.727}/\textbf{0.723} & \textbf{1.25}/\textbf{1.50} & \textbf{0.693}/0.677 & \textbf{2.00}/\textbf{2.00}\\
$\methodname_{\text{LPIPS}_{2}}$ (ours) & 0.753/0.743 & 2.00/2.25 & \textbf{0.736}/0.724 & 1.75/2.00 & 0.690/0.676 & 2.25/2.25 & 0.692/\textbf{0.678} & \textbf{2.00}/\textbf{2.00}\\

\bottomrule
\end{tabular}}
\end{table*}

The newly suggested score function is designed to capture infinitesimal distortions of reconstructed images sampled from the \textbf{test dataset} through the lens of autoencoders $(\ie, \text{AE}(\rvx))$ pre-trained on the \textbf{train dataset} $\nu$ (\eg, LAION-aesthetics \citep{Schumann22LAION}). By taking a directional derivative in the direction of $\rvx - \mathcal{F}(\rvx)$, the score function amplifies the difference (\ie, distribution shift) between the train and test dataset in high-frequency information. We note that this further reduces the dependency on the background since low-frequency information is not taken into account for the score function.

On the other hand, on text-to-image generative models that are trained on large-scale data distribution similar to $\nu$, we expect the HFI score to be lower on their generated data. Furthermore, our design of Eq \ref{eq:sungwoo_HFI} is also motivated by architectural similarities of generative models. Generative models employ upsampling layers composed of a series of convolution kernels \citep{Rombach22SD,Dhariwal21ADM}. We anticipate that our spatial term, $\rvx-\mathcal{F}(\rvx)$, will capture these local correlations, resulting in lower HFI scores.

Unfortunately, the numerical estimation of the gradient in Eq~\ref{eq:sungwoo_HFI} might be challenging since both distance function components depend on the input. Hence, we estimate the numerical approximation of \methodname by taking the $1$st-order Taylor series expansion.

\begin{align}\label{eq:eq2}
    \left\langle \frac{\partial d(\rvx,\text{AE}(\nu, \rvx))}{\partial \rvx} \; , \; \rvx-\mathcal{F}(\rvx)  \right\rangle     
   & \\ ~\approx~  d(\rvx,\text{AE}(\rvx))  - & d(\mathcal{F}(\rvx), \text{AE}(\mathcal{F}(\rvx))) \nonumber  
\end{align}


 We expect \methodname can effectively discriminate the real data (\ie, $ \methodname (\rvx) > \tau$) or generated data  (\ie, $ \methodname(\rvx) \le \tau$).  We verify our hypothesis on the same toy dataset. For consistency, we use LPIPS for the distance function $d$. Since the optimal filter choice is unknown, a Gaussian kernel with kernel size $k=3$ and standard deviation $\sigma=0.8$ is selected for the experiment as a de facto choice. The choice of $k=3$ is informed by the architecture of the autoencoders in LDMs, which utilize 3$\times$3 kernels. 
 

 We report the results in Figure \ref{fig:toy:hfi}. \methodname successfully distinguishes real image distribution from the AI-generated data. \methodname also successfully distinguishes SDv1.5-generated data from other AI-generated data (\eg, SDv2-base, Kandinsky).

When the autoencoders from distinct models, $\text{AE}_{1},...,\text{AE}_{n}$, are available, we follow the practice of \citep{ricker24aeroblade} and finalize the ensemble version of \methodname as follows.

\begin{equation}\label{eq:eq3}
    \text{HFI}_{d,\mathcal{F}}(\rvx) = \min_{i} \text{HFI}_{d,\mathcal{F},\text{AE}_{i},\nu_{i}}(\rvx)
\end{equation}


Notably, the ensemble approach aligns with real-world scenarios, where the optimal autoencoder for a given dataset is typically unknown in advance.

\section{Experiment}
\label{sec:exp}

In this section, we evaluate the efficacy of our proposed \methodname in detecting images generated by text-to-image generative models. First, we introduce our experiment protocols and the datasets (Section \ref{ssec:expsetup}). We then present our main results (Section \ref{ssec:mainresults}) followed by extensive ablation studies (Section \ref{ssec:ablationstudy}).

\subsection{Experimental setups}
\label{ssec:expsetup}

\begin{table*}[t]
\centering
\caption{AI-generated image detection performance (AUPR) of \methodname and baselines in the GenImage \citep{Zhu23genimage} dataset. \textbf{Bold} and \underline{underline} denotes the best and second best methods.}
\label{table:ensemble_genimage}
\resizebox{\textwidth}{!}{
\begin{tabular}{cccccccccc}
\toprule
                                 
Method  & ADM & BigGAN & GLIDE & Midjourney & SD1.4 & SD1.5 & VQDM & Wukong & Mean \\ 
\midrule 
\multicolumn{10}{c}{\emph{Training-based Detection Methods} } \\ 
\midrule
DRCT/UnivFD \citep{Chen24DRCT} & 0.892 & 0.924 & 0.964 & 0.974 & 0.997 & 0.995 & 0.966 & 0.994 & 0.963\\
NPR \citep{Tan24NPR} &   0.733 & 0.920 & 0.924 & 0.822 & 0.842 & 0.841 & 0.766 & 0.814 & 0.833\\
\midrule
\multicolumn{10}{c}{\emph{Training-free Detection Methods}} \\ 
\midrule
RIGID \citep{He24rigid}   & 0.790 & 0.976 & 0.964 & 0.797 & 0.698 & 0.699 & 0.860 & 0.708 & 0.812\\

$\text{AEROBLADE}_{\text{LPIPS}}$  & 0.748 & 0.919 & 0.977 & 0.984 & 0.974 & 0.976 & 0.596 & 0.977 & 0.894 \\
$\text{AEROBLADE}_{\text{LPIPS}_{2}}$ \citep{ricker24aeroblade} & 0.838 & 0.986 & 0.990 & 0.988 & 0.983 & 0.984 & 0.723 & 0.984 & 0.935\\
$\methodname_{\text{LPIPS}}$ \textbf{(ours)} & \underline{0.872} & \underline{0.989} & \underline{0.994} & \underline{0.996} & \textbf{0.998} & \textbf{0.998} & \underline{0.884} & \underline{0.997} & \underline{0.966} \\
$\methodname_{\text{LPIPS}_{2}}$ \textbf{(ours)} & \textbf{0.923} & \textbf{0.996} & \textbf{0.995} & \textbf{0.998} & \textbf{0.998} & \textbf{0.998} & \textbf{0.905} & \textbf{0.999} & \textbf{0.977}\\
\bottomrule
\end{tabular}}
\end{table*}

\begin{table*}[t]
\centering
\caption{AI-generated image detection performance (AUPR) of \methodname and baselines in the SynthBuster dataset. \textbf{Bold} denotes the best method.}
\label{table:ensemble_synthbuster}
\resizebox{\textwidth}{!}{
\begin{tabular}{ccccccccccc}
\toprule
                                 
Method  & Firefly & GLIDE & SDXL & SDv2 & SDv1.3 & SDv1.4 & DALL-E 3 & DALL-E 2 & Midjourney & Mean \\ 

\midrule
RIGID \citep{He24rigid}   & 0.543 & 0.866 & 0.779 & 0.593 & 0.473 & 0.472 & 0.470 & \textbf{0.596} & 0.614 & 0.601\\

$\text{AEROBLADE}_{\text{LPIPS}}$  & 0.617 & 0.938 & 0.734 & 0.914 & 0.938 & 0.940 & 0.638 & 0.458 & 0.952 & 0.792 \\
$\text{AEROBLADE}_{\text{LPIPS}_{2}}$ \citep{ricker24aeroblade} & 0.584 & 0.964 & 0.755 & 0.959 & 0.967 & 0.968 & 0.579 & 0.418 & 0.969 & 0.796\\
$\methodname_{\text{LPIPS}}$ & \textbf{0.692} & 0.981 & 0.689 & 0.992 & 0.996 & 0.996 & 0.532 & 0.516 & 0.992 & 0.821 \\
$\methodname_{\text{LPIPS}_{2}}$ & 0.581 & \textbf{0.986} & \textbf{0.796} & \textbf{0.996} & \textbf{0.999} & \textbf{0.999} & \textbf{0.648} & 0.425 & \textbf{0.997} & \textbf{0.825}\\
\bottomrule
\end{tabular}}
\end{table*}

\begin{table}[t]
\centering
\caption{AI-generated image detection performance (AUPR) of \methodname and training-free baselines in the DiffusionFace \citep{Chen24diffusionface} dataset. \textbf{Bold} and \underline{underline} denotes the best and second best methods.}
\label{table:ensemble_diffusionface}
\resizebox{0.48\textwidth}{!}{
\begin{tabular}{cccccc}
\toprule
                                 
Method  & SDv1.5 T2I & SDv2.1 T2I & SDv1.5 I2I & SDv2.1 I2I & Mean \\ 
\midrule 

RIGID &  0.620 & 0.531 & 0.529 & 0.542 & 0.556 \\

$\text{AEROBLADE}_{\text{LPIPS}}$ & 0.832 & 0.486 & 0.910 & \textbf{0.613} & 0.710 \\
$\text{AEROBLADE}_{\text{LPIPS}_{2}}$ & 0.813 & 0.459 & \underline{0.964} & 0.579 & 0.704\\
$\methodname_{\text{LPIPS}}$ (ours) & \textbf{0.878} & \textbf{0.584} & 0.942 & \underline{0.599} & \textbf{0.751}\\
$\methodname_{\text{LPIPS}_{2}}$ (ours) & \underline{0.853} & \underline{0.536} & \textbf{0.985} & 0.592 & \underline{0.742} \\
\bottomrule
\end{tabular}}
\vskip -0.1in
\end{table}

\begin{figure*}[t]
\centering

\begin{subfigure}[b]{0.47\textwidth}
\includegraphics[width=0.96\textwidth]{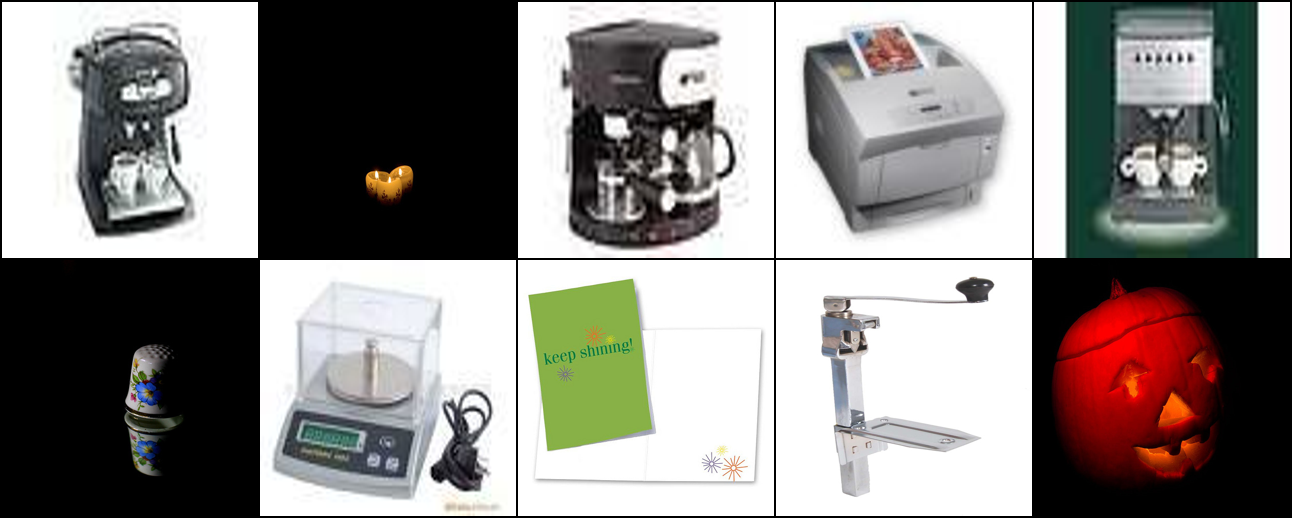}
\caption{Real images with low reconstruction error}\label{fig:viz:real}
\end{subfigure}
\hfill
\begin{subfigure}[b]{0.47\textwidth}
\includegraphics[width=0.96\textwidth]{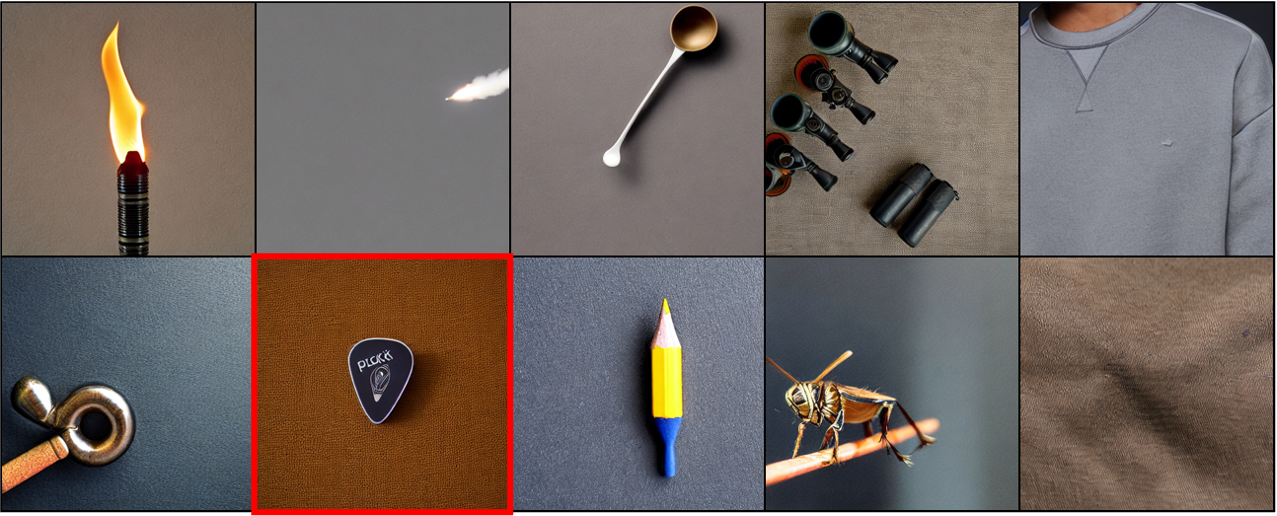}
\caption{Generated images with high reconstruction error}\label{fig:viz:fake}
\end{subfigure}
\caption{\textbf{Visualization of the edge-cases.} \textbf{(a)} Visualization of the ImageNet data where AEROBLADE outputs the smallest uncertainty. \textbf{(b)} Visualization of the SDv1.4-generated data where AEROBLADE outputs the highest uncertainty. We mark the sample where \methodname also fails.}
\label{fig:visualize_error}
\end{figure*}

\noindent
\textbf{Evaluation protocols.} Our main experiments are divided into two scenarios. We first report the performance between the autoencoder-based methods in the cross-autoencoder setup. To be specific, we compare \methodname against the autoencoder-based baseline, AEROBLADE \citep{ricker24aeroblade}, under various autoencoder choices.  We use the autoencoder of the Stable diffusion (SD)v1.4/v2-base \citep{Rombach22SD}, Kandinsky2.1 \citep{Razzhigaev23Kandinsky}, and MiniSD-diffusers \citep{MiniSD} throughout the experiment. Then, we compute the ensemble performance of \methodname and AEROBLADE and report their performance against the other AI-generated image detection methods. We mainly report the area of the region under the Precision-Recall curve (AUPR) and the area of the region under the ROC curve (AUROC). We additionally report the average rank across the dataset in the cross-autoencoder setup. \\ 

\noindent
\textbf{Datasets.} We test the efficacy of \methodname applied in various data domains. For the natural image domain, we experiment on the GenImage \citep{Zhu23genimage} benchmark where the real image is from the ImageNet \citep{Deng09ImageNet}. GenImage constitutes data generated by 8 generative models that are not limited to LDMs: ADM \citep{Dhariwal21ADM}, BigGAN \citep{Brock18Biggan}, GLIDE \citep{Nichol22GLIDE}, Midjourney \citep{22Midjourney}, SD v1.4/1.5 \citep{Rombach22SD}, VQDM \citep{Gu22VQDM}, and Wukong \citep{Gu22Wukong}. We also evaluate \methodname on the Synthbuster \citep{QuentinSynthbuster} benchmark where the real image is from the Raise-1k \citep{Raise2015Nguyen} dataset. The benchmark contains 9 text-to-image generative models: Firefly, GLIDE \citep{Nichol22GLIDE}, SD v1.3/1.4/2/XL \citep{Rombach22SD}, DALL-E 2/3 \citep{ramesh2022dalle2}, and Midjourney \citep{22Midjourney}. Finally, we evaluate \methodname on the DiffusionFace \citep{Chen24diffusionface} benchmark where the real image is from the Multi-Modal-CelebA-HQ \citep{Xia21MultiCelebA} dataset. The benchmark contains 4 LDM-generated categories: images generated from the noise via SDv1.5/2.1 models (SD1.5 T2I, SD2.1 T2I) and images generated from image-to-image translation starting from real images via SDv1.5/2.1 models (SDv1.5 I2I, SDv2.1 I2I).\\

\noindent
\textbf{Baselines.} 
First, we compare \methodname against AEROBLADE \citep{ricker24aeroblade} in the cross-autoencoder setup. Since original AEROBLADE uses the variant of LPIPS \citep{zhang18LPIPS} that is based on the 2nd layer of VGG \citep{simonyan15VGG} network, we denote it as $\text{AEROBLADE}_{\text{LPIPS}_{2}}$ and also experiment our \methodname on the distance, denoted as $\text{\methodname}_{\text{LPIPS}_{2}}$. We also report the performance of AEROBLADE based on the full LPIPS distance, denoted as $\text{AEROBLADE}_{\text{LPIPS}}$. In the second scenario, we also report the performance of RIGID \citep{He24rigid}, a recently proposed training-free detection method. We also report the performance of competitive training-based methods: NPR \citep{Tan24NPR} and DRCT \citep{Chen24DRCT}. Both methods are trained on ImageNet data and SDv1.4-generated data. \\

\noindent
\textbf{Design choices and implementation details.}
All experiments on \methodname are done in fixed choices of the distance function $d$, the low-pass filter $\mathcal{F}$, and the ensemble strategy. For the distance function $d$, we use the standard LPIPS distance and that of AEROBLADE and denote the performance as $\text{HFDD}_{\text{LPIPS}}$ and $\text{HFDD}_{\text{LPIPS}_{2}}$, respectively. For the low pass-filter $\mathcal{F}$, We use a 3 $\times$ 3 Gaussian filter with a standard deviation of 0.8 consistent with Section \ref{sec:method}. 

We crop and resize all images to fit the default dimension of the autoencoder (\ie, 256 $\times$ 256 in MiniSD \citep{MiniSD}, and 512 $\times$ 512 for the rest). Specifically, we crop the dimension of the image that exceeds the autoencoder dimension and resize the image that is smaller than the autoencoder dimension. We implement the code on the PyTorch \citep{Paskze19Pytorch} framework. We refer to the Appendix for further details.

\subsection{Main results}
\label{ssec:mainresults}

Table \ref{table:crossae_genimage_main} and Table \ref{table:crossae_diffusionface_main} report the performance of \methodname and AEROBLADE in the cross-autoencoder setup on the GenImage and DiffusionFace dataset, respectively. We refer to 
 the Appendix for the full results. \methodname outperforms AEROBLADE on all benchmarks and autoencoder choices on average. It is noteworthy that \methodname outperforms AEROBLADE in 61 out of 64 experiments and 26 out of 32 on the GenImage and DiffusionFace benchmarks, respectively.

Table \ref{table:ensemble_genimage}, \ref{table:ensemble_synthbuster}, and \ref{table:ensemble_diffusionface} report the ensemble performance of \methodname compared to various AI-generated image detection methods. \methodname achieves the best performance on most generative models examined. It is worth noting that HFI shows major improvements even when the model is not from the examined SD models (\eg ADM, VQDM in Table \ref{table:ensemble_genimage}). Furthermore, HFI achieves competitive performance to DRCT \citep{Chen24DRCT} and NPR \citep{Tan24NPR}, outperforming in 7 and 8 out of 8 experiments. Finally, HFI achieves near-perfect detection performance in benchmarks where the underlying autoencoder is available for computation (\eg SD-based, Wukong, Midjourney). 


We further analyze the failure cases of AEROBLADE and \methodname where the underlying autoencoder of the generated data is the same as the autoencoder used in detection. Figure \ref{fig:viz:real} and \ref{fig:viz:fake} report the 10 real samples and 10 SDv1.4-generated samples where AEROBLADE  outputs the lowest and highest uncertainty with the average of 0.003 and 0.029, respectively. Aligned to the results on the toy dataset in Section \ref{ssec:motivation}, AEROBLADE fails to detect real data with simple backgrounds. While  \methodname also shows low uncertainty in these samples, \methodname successfully distinguishes most cases by lowering the average score of the generated data in Figure \ref{fig:viz:fake} $\ie, 0.029 \rightarrow -0.011$.

\subsection{Ablation studies}
\label{ssec:ablationstudy}

\begin{table}[t]
\centering
\caption{Mean AI-generated image detection performance (AUPR) of \methodname to the kernel size $k$ and standard deviation $\sigma$ of the Gaussian filter $\mathcal{F}$. We report the average performance computed across the GenImage dataset. \textbf{Bold} denotes the best hyperparameter.}
\label{table:ablation_gaussian}
\resizebox{0.48\textwidth}{!}{
\begin{tabular}{ccccccccccc}
\toprule
 & \multicolumn{3}{c}{AE: SDv2-base} & \multicolumn{3}{c}{AE: Kandinsky} &  \multicolumn{3}{c}{AE: MiniSD}\\ \cmidrule(lr){2-4}\cmidrule(lr){5-7}\cmidrule{8-10}  
  & $k=3$ & $k=5$ & $k=7$ & $k=3$ & $k=5$ & $k=7$ & $k=3$ & $k=5$ & $k=7$ \\ 
\midrule
$\sigma = 0.5$ & 0.863 & 0.863 & 0.863 & 0.893 & 0.894 & 0.894 & 0.897 & 0.896 & 0.896\\
$\sigma = 0.8$ & 0.917 & 0.916 & 0.916 & \textbf{0.900} & 0.895 & 0.894 & 0.925 & 0.922 & 0.922\\
$\sigma = 1.1$ & \textbf{0.922} & 0.919 & 0.917 & 0.897 & 0.886 & 0.882 & \textbf{0.931} & 0.921 & 0.917\\
$\sigma = 1.4$ & \textbf{0.922} & 0.918 & 0.913 & 0.895 & 0.878 & 0.869 & \textbf{0.931} & 0.915 & 0.905\\

\bottomrule
\end{tabular}}
\vskip -0.1in
\end{table}

\begin{table}[t]
\centering
\caption{Ablation studies of \methodname under different low-pass filter. We report the ensemble performance. \textbf{Bold} denotes the best mapping choice.}
\label{table:ablation_filter}
\resizebox{0.48\textwidth}{!}{
\begin{tabular}{cccccccccc}
\toprule
                                 
$\mathcal{F}$ & ADM & BigGAN & GLIDE & Midj & SD1.4 & SD1.5 & VQDM & Wukong & Mean \\ 
\midrule 
Gaussian blur & 0.872 & \textbf{0.989} & \textbf{0.994} & \textbf{0.996} & \textbf{0.998} & 0.998 & \textbf{0.884} & 0.997 & \textbf{0.966} \\
Box blur & \textbf{0.873} & 0.987 & 0.993 & \textbf{0.996} & \textbf{0.998} & \textbf{0.999} & 0.868 & \textbf{0.998} & 0.964 \\
Bilateral blur & 0.778 & 0.918 & 0.928 & 0.962 & 0.975 & 0.975 & 0.694 & 0.979 & 0.901\\
Median blur & 0.865 & 0.972 & 0.971 & 0.988 & 0.996 & 0.996 & 0.853 & 0.996 & 0.954\\
DCT & 0.709 & 0.888 & 0.969 & 0.960 & 0.976 & 0.977 & 0.632 & 0.984 & 0.887  \\

\bottomrule
\end{tabular}}
\end{table}

\begin{table}[t]
\centering
\caption{Ablation studies of \methodname under different distance functions. We report the ensemble performance. \textbf{Bold} denotes the best distance choice.}
\label{table:ablation_distance}
\resizebox{0.48\textwidth}{!}{
\begin{tabular}{cccccccccc}
\toprule
                                 
$d$ & ADM & BigGAN & GLIDE & Midj & SD1.4 & SD1.5 & VQDM & Wukong & Mean \\ 
\midrule 
LPIPS & 0.872 & 0.989 & 0.994 & 0.996 & \textbf{0.998} & \textbf{0.998} & 0.884 & 0.997 & 0.966\\
$\text{LPIPS}_{1}$ & 0.910 & 0.994 & \textbf{0.995} & 0.970 & 0.986 & 0.987 & 0.901 & 0.986 & 0.966\\
$\text{LPIPS}_{2}$  & \textbf{0.923} & \textbf{0.996} & \textbf{0.995} & \textbf{0.998} & \textbf{0.998} & \textbf{0.998} & \textbf{0.905} & \textbf{0.999} & \textbf{0.977}\\
$\text{LPIPS}_{3}$ & 0.887 & 0.990 & 0.992 & 0.991 & 0.994 & 0.995 & 0.847 & 0.997 & 0.962\\
$\text{LPIPS}_{4}$ & 0.763 & 0.689 & 0.873 & 0.910 & 0.940 & 0.939 & 0.722 & 0.936 & 0.847\\
$\text{LPIPS}_{5}$ & 0.682 & 0.632 & 0.774 & 0.822 & 0.860 & 0.853 & 0.641 & 0.839 & 0.763\\
$\text{DISTS}$ & 0.678 & 0.823 & 0.871 & 0.662 & 0.751 & 0.751 & 0.551 & 0.720 & 0.726\\
\bottomrule
\end{tabular}}

\end{table}

\noindent
\textbf{Design choices.} We extensively explore design choices for \methodname under the GenImage dataset. First, we perform a hyperparameter analysis on the size of the kernel $k$ and the standard deviation $\sigma$ for the Gaussian blur filter $\mathcal{F}$ under different autoencoders. We report the result in Table \ref{table:ablation_gaussian}. Note that $k=3, \sigma=0.8$ refers to the original setting. $k=3$ performs the best consistently. 

We further explore different choices of low-pass filtering. We apply box, bilateral, and median blur filters of kernel size 3 for the experiment. We also experiment with a discrete cosine transform (DCT) based frequency nulling scheme that erases high-frequency components above a given frequency. Since the size of the hyperparameter search space on the DCT is massive, we instead rely on our rule of thumb based on antialiasing. Namely, we view the encoder as a downsampling kernel and propose to set the cutoff frequency as the Nyquist frequency. We follow the practice of \citet{Chen24Aliasing} for calculating the Nyquist frequency.  

We report the result in Table \ref{table:ablation_filter}. Surprisingly, the box filter also shows competitive performance, slightly outperforming the Gaussian filter on average. On the other hand, our designed hyperparameter on DCT  underperforms over AEROBLADE. We show further experiment results in the Appendix.

\begin{figure}[t]
\centering

\begin{subfigure}[b]{0.23\textwidth}
\includegraphics[width=\textwidth]{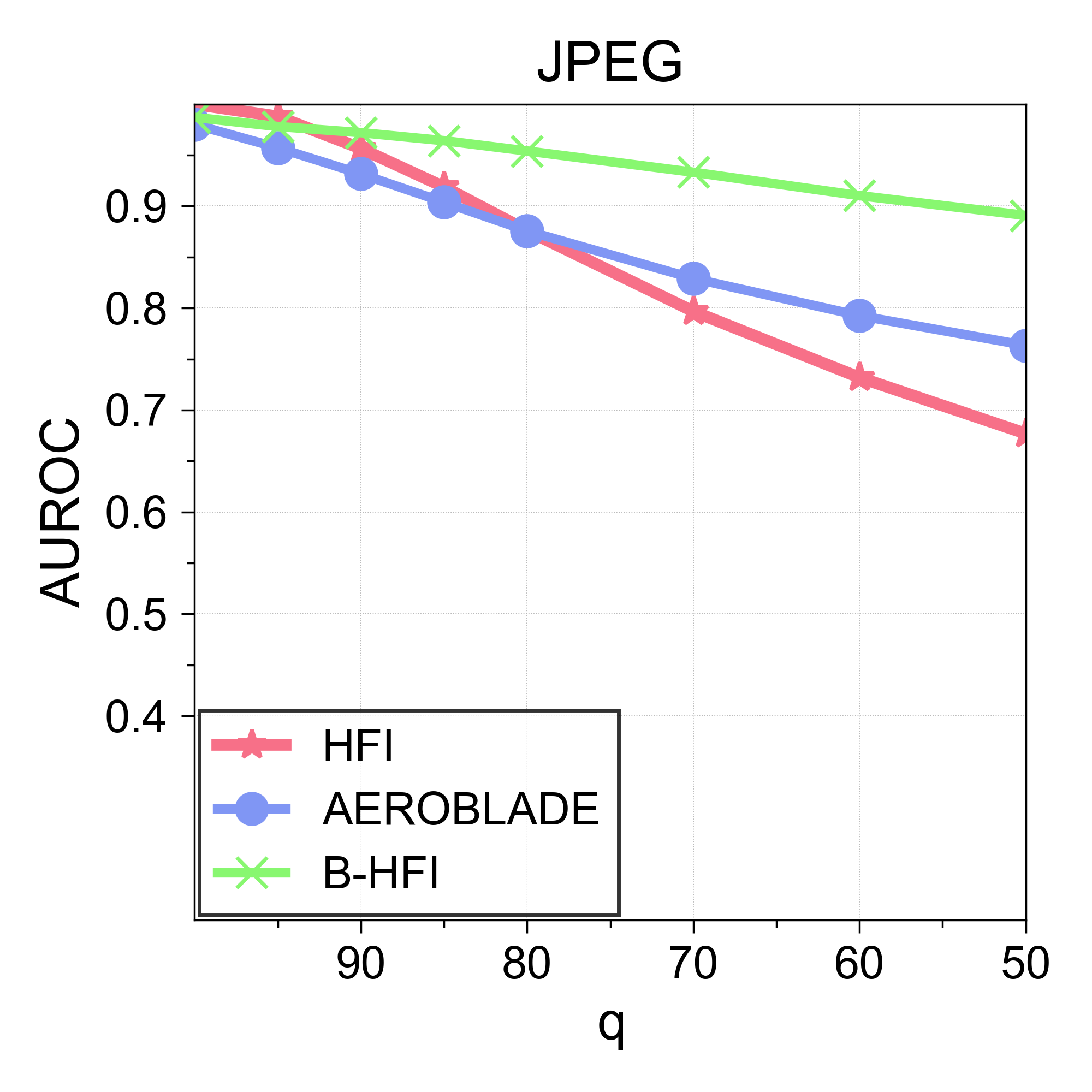}
\caption{JPEG}\label{fig:cor:jpeg}
\end{subfigure}
\hfill
\begin{subfigure}[b]{0.23\textwidth}
\includegraphics[width=\textwidth]{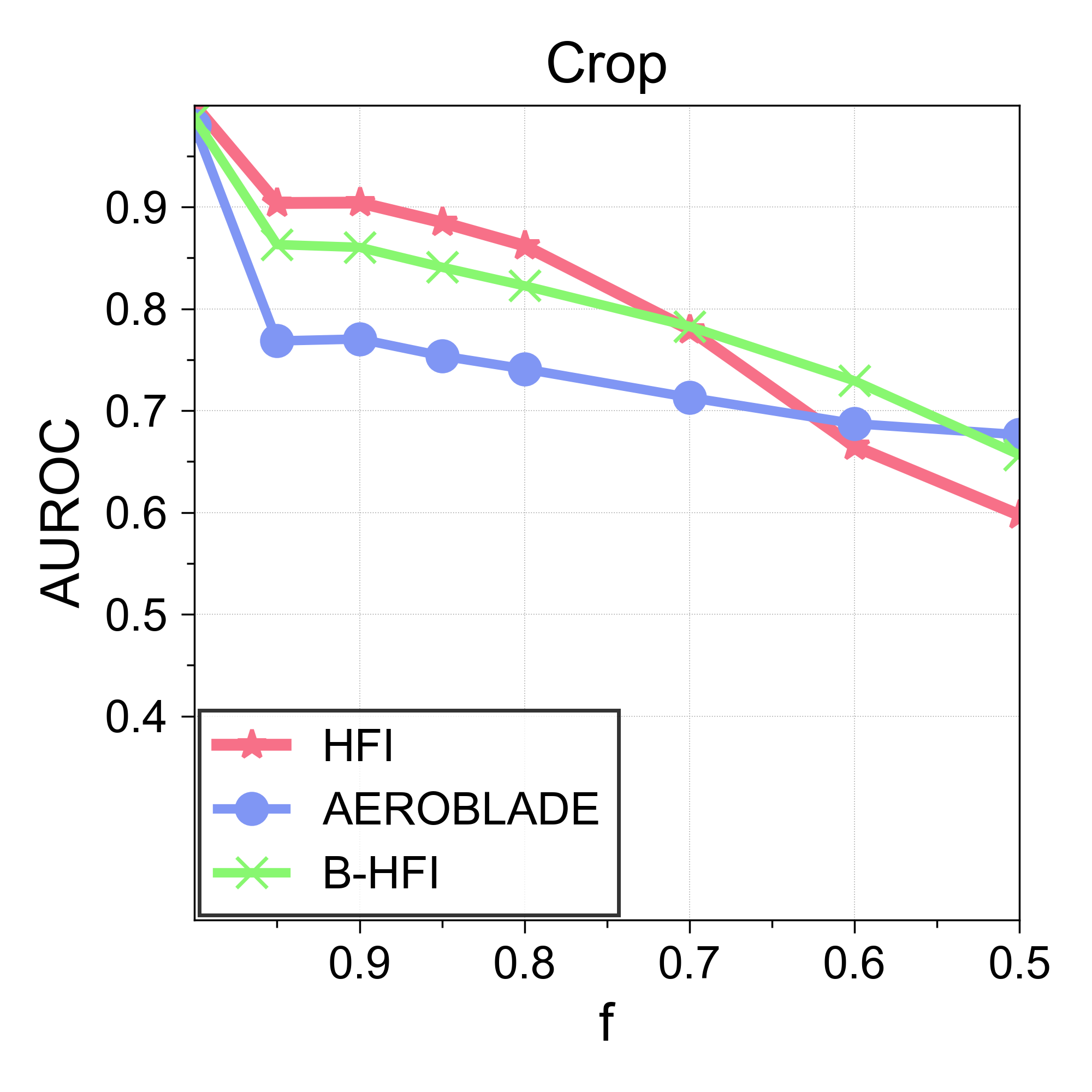}
\caption{Crop}\label{fig:cor:crop}
\end{subfigure}
\hfill
\caption{Performance of \methodname, AEROBLADE, and B-HFI under corruption.}
\label{fig:corruption}
\end{figure}

We also experiment with different choices of perceptual distance. For the alternative distance, we experiment $\text{LPIPS}_{1}$,$\text{LPIPS}_{3}$,$\text{LPIPS}_{4}$, $\text{LPIPS}_{5}$, and DISTS \citep{Ding20DISTS}. We report the ensemble performance consistent with the results of Table \ref{table:ensemble_genimage}. We report the results in Table \ref{table:ablation_distance}. $\text{LPIPS}_{2}$ shows the best performance. \\

\begin{table*}[t]
\centering
\caption{Belonging vs non-belonging image detection performance (AUPR) of \methodname compared to LatentTracer \citep{Wang24LatentTracer}. We denote $\mathcal{M}_{1}$  and $\mathcal{M}_{2}$ to the belonging and non-belonging model, respectively. \textbf{Bold} denotes the best method.}
\label{table:latenttracer}
\resizebox{\textwidth}{!}{
\begin{tabular}{ccccccccccc}
\toprule
 & \multicolumn{3}{c}{$\mathcal{M}_{1}$: SDv1.5} & \multicolumn{3}{c}{$\mathcal{M}_{1}$: SDv2-base} & \multicolumn{3}{c}{$\mathcal{M}_{1}$: Kandinsky } \\ \cmidrule(lr){2-4}\cmidrule(lr){5-7}\cmidrule(lr){8-10}
Method  & SDv2-base & SDv2.1 & Kand & SDv1.5 & SDv2.1 & Kand & SDv1.5 & SDv2-base& SDv2.1\\ 
\midrule 

LatentTracer & 0.9990 & 0.9983 & \textbf{0.9976} & \textbf{0.9994} & 0.9990 & 0.9984  & 0.9973 & 0.9971 & 0.9945\\
$\text{AEROBLADE}_{\text{LPIPS}_{2}}$ & 0.9652 & 0.9935 & 0.8145 & 0.9392 &  0.9941 & 0.8271 & 0.9950 & 0.9982 & 0.9987 \\
$\methodname_{\text{LPIPS}_{2}}$ (ours) & \textbf{0.9999}& \textbf{1.0000} & 0.9972 & 0.9985 & \textbf{0.9998} & \textbf{0.9989} & \textbf{0.9986} & \textbf{0.9986} & \textbf{0.9988}\\
\bottomrule
\end{tabular}}
\end{table*}

\noindent
\textbf{Robustness to corruption.} While our \methodname has achieved great success on various benchmarks, the underlying high-frequency information can be affected by corruption. For example, applying JPEG \citep{Wallace92JPEG} transform to the image generates high-frequency artifacts to the corrupted image. To better understand our limitations on severe corruptions, we test \methodname when the AI-generated and real data are corrupted. We experiment when JPEG compression, and crop are applied for corruption. We test \methodname and AEROBLADE on the ImageNet vs SDv1.4 task with the autoencoder of SDv1.4 as the $\text{AE}$.

We report the result in Figure \ref{fig:corruption}. Our method is relatively robust in small corruptions with larger gaps against AEROBLADE than when evaluated in clean images (\eg, Crop (Figure \ref{fig:cor:crop})). However, under severe corruption, HFI's performance deteriorates.

Since low-pass filtering has been established as a standard practice for mitigating the effects of external noise, we also experiment with $\text{B-HFI}(\rvx) = \text{HFI}(\mathcal{F}_{\text{B}}(\rvx))$. We use a Gaussian filter with $k=3$ and $\sigma=0.8$ for $\mathcal{F}_{\text{B}}$. We also report their result in Figure \ref{fig:corruption} denoted as B-HFI. The B-HFI method enhances robustness compared to HFI while outperforming AEROBLADE on the clean data.
\section{Tracing LDM-generated images}
\label{sec:ldmtrace}

We further elaborate on tracing LDM-generated images introduced in Section \ref{ssec:ltracebasic}. To be specific, for a given LDM model $\mathcal{M}_{1}$ and its autoencoder $\text{AE}_{\mathcal{M}_{1}}$, we directly apply $\text{\methodname}_{\text{AE}_{\mathcal{M}_{1}}}$ to distinguish the belonging images generated from $\mathcal{M}_{1}$ from the others. Namely, we regard the input image $\mathbf{x}$ as generated from the model $\mathcal{M}_{1}$ if $\text{\methodname}_{\text{AE}_{\mathcal{M}_{1}}}$ outputs low score.

We follow the practice of \citet{Wang24LatentTracer} and consider the case where both $\mathcal{M}_{1}$ and $\mathcal{M}_{2}$ are from 4 different LDMs: SDv1.5, SDv2-base, SDv2.1, and Kandinsky2.1. For experiments, we follow the setup of \citet{Wang24LatentTracer} where they sample 54 prompts and generate 10 images per prompt, resulting in 540 images per model. We also compare against LatentTracer \citep{Wang24LatentTracer}, the best baseline that performs input optimization in the test time. We test LatentTracer on the official code released by the authors. We report the performance of $\text{\methodname}_{\text{LPIPS}_{2}}$ and $\text{AEROBLADE}_{\text{LPIPS}_{2}}$.

We report the result in Table \ref{table:latenttracer}. Both \methodname and LatentTracer achieve near-perfect detection performance. On the other hand, AEROBLADE struggles in "$\mathtt{SD \Hquad vs \Hquad Kandinsky}$" tasks. It is worth noting that \methodname is much more efficient in time complexity than LatentTracer. Namely, \methodname takes 0.255s/sample, which achieves 57x speedup against the LatentTracer which takes 14.65s/sample in A100 gpus.
\section{Related Works}
\label{sec:related}

\textbf{Training-based AI-generated image detection.} Most training-based AI-generated image detection methods follow a two-stage setup: the key representation is extracted from the image followed by training a binary classifier over the key representations. A variety of representations have been proposed, including handcrafted spatial features \citep{Popescu05Handcraft},  patch statistics \citep{Chai20Patch}, a feature of the foundation model \citep{Ohja23UnivFD}, or their unification \citep{Yan24Combine}.

Leveraging frequency maps as a key representation has also been a major practice of training-based AI-generated image detection methods. \citet{frank20GANfreq} discover that the GAN-generated images contain noticeable frequency artifacts. \citet{Li21freq} fuse the spatial and frequency representation and adopt a single-center loss for pulling out manipulated face representation. \citet{Wang23freq} introduce a relation module that connects multiple domains. \citet{Tan24AAAIFreq} incorporate the FFT module inside the classifier. While our method also utilizes high-frequency information from the data, we directly incorporate the reconstruction distance for examination instead of training on the extracted features.

\noindent
\textbf{Training-free AI-generated image detection.} 
Since no training data is available, training-free AI-generated image detection methods utilize a pre-trained model to design the uncertainty function. \citet{ma2023exposing} utilize the error of deterministic forward-reverse diffusion process to detect diffusion-generated images. \citet{ricker24aeroblade} utilize the perceptual distance between the input and its reconstruction through the autoencoder of LDMs to detect LDM-generated images. \citet{He24rigid} utilizes the cosine similarity on the feature of Dinov2 \citep{oquab24dinov2} between the input data and its perturbation. The perturbation is constructed by injecting Gaussian Noise into the input. 


\noindent
{\bf Aliasing in image domain. } \citet{Zhang19Convolutionshift} discusses aliasing in the classification task that aliasing occurs while in subsampling and designs a blurring-based pooling operation for the remedy. \citet{Vasconcelos21antialiasing} propose the variant by positioning the blurring operation before the non-linearities. \citet{Karras21upsamplingaliasing} propose a custom upsampling layer to deal with aliasing while training generative adversarial networks. \citet{agnihotri24antialiasing} also modify the operation in the upsampling layer in the pixel-wise prediction task. However, to our knowledge, most of the existing works focus on the training phase. \\


\section{Conclusion}
\label{sec:conclusion}

This paper proposes \methodname, a novel, training-free AI-generated image detection method based on the autoencoder of LDM. \methodname measures the influence of the high-frequency spatial component on the distortion of the reconstructed image through the autoencoder. \methodname can effectively distinguish 3 types of distributions: images generated from the given autoencoder, other AI-generated images, and real images. Hence, \methodname achieves the state-of-the-art in both training-free AI-generated image detection and generated image detection of the specified model. 

While the proposed work mainly discusses the detection of LDM-generated images as a representative of training-free detection, some of the text-to-image generative models do not utilize the explicit autoencoder architecture (\eg BigGAN, ADM). Furthermore, explicit access to the LDM autoencoder may not be available in the proprietary model (\eg Firefly, Dall-E 3). While our method is competitive against baselines in detecting images generated by such a model, applying representations from other foundation models might be efficient. We leave this direction for future work.

\bibliography{cvpr}
\bibliographystyle{iclr2025_conference}

\newpage
\appendix
\appendix

\section{Full results on the cross-autoencoder setup}

\begin{table*}[t]
\centering
\caption{Cross-autoencoder AI-generated image detection performance (AUROC/AUPR) of \methodname and AEROBLADE \citep{ricker24aeroblade} in the GenImage \citep{Zhu23genimage} dataset. \textbf{Bold} denotes the best method.}
\label{table:crossae_genimage_appendix}
\resizebox{\textwidth}{!}{
\begin{tabular}{cccccccccc}
\toprule
                                 
Method  & ADM & BigGAN & GLIDE & Midj & SD1.4 & SD1.5 & VQDM & Wukong & Mean \\ 
\midrule 
 \multicolumn{10}{c}{\emph{AE: SDv1.4 \citep{Rombach22SD}} } \\ 
\midrule
$\text{AEROBLADE}_{\text{LPIPS}}$  & 0.804/0.757 & 0.889/0.909 & 0.975/0.976 & \textbf{0.921}/\textbf{0.928} & 0.980/0.986 & 0.981/0.986 & 0.640/0.595 & 0.983/0.988 & 0.897/0.891\\
$\text{AEROBLADE}_{\text{LPIPS}_{2}}$ & 0.856/0.833 & 0.981/0.987 & 0.989/0.990 & 0.918/\textbf{0.928} & 0.982/0.988 & 0.984/0.989 & 0.732/0.712 & 0.983/0.989 & 0.928/0.927\\
$\methodname_{\text{LPIPS}}$ (ours) &  0.886/0.857 & 0.992/0.994 & \textbf{0.996}/\textbf{0.996} & 0.898/0.883 & \textbf{1.000}/\textbf{1.000} & \textbf{0.999}/\textbf{1.000} & 0.794/0.770 & \textbf{0.999}/\textbf{0.999} & 0.946/0.937\\
$\methodname_{\text{LPIPS}_{2}}$ (ours) & \textbf{0.918}/\textbf{0.923} & \textbf{0.994}/\textbf{0.997} & 0.995/\textbf{0.996} & 0.907/0.915 & 0.999/0.999 & \textbf{0.999}/0.999 & \textbf{0.862}/\textbf{0.888} & \textbf{0.999}/\textbf{0.999} & \textbf{0.959}/\textbf{0.965}\\

\midrule
 \multicolumn{10}{c}{\emph{AE: SDv2-base \citep{Rombach22SD}}} \\ 
\midrule
$\text{AEROBLADE}_{\text{LPIPS}}$  & 0.800/0.753 & 0.905/0.922 & 0.976/0.977 & 0.981/0.984 & 0.819/0.848 & 0.823/0.850 & 0.627/0.586 & 0.838/0.864 & 0.846/0.848  \\
$\text{AEROBLADE}_{\text{LPIPS}_{2}}$ & 0.856/0.834 & 0.979/0.986 & 0.988/0.989 & 0.985/0.989 & 0.807/0.842 & 0.810/0.842 & 0.725/0.710 & 0.833/0.865 & 0.873/0.882 \\
$\methodname_{\text{LPIPS}}$ (ours) &  0.883/0.859 & 0.987/0.991 & 0.993/0.994 & 0.996/0.997 & \textbf{0.885}/\textbf{0.903} & \textbf{0.891}/\textbf{0.906} & 0.810/0.778 & 0.888/0.907 & 0.916/0.917\\
$\methodname_{\text{LPIPS}_{2}}$ (ours) & \textbf{0.921}/\textbf{0.929} & \textbf{0.993}/\textbf{0.996} & \textbf{0.995}/\textbf{0.996} & \textbf{0.998}/\textbf{0.998} & 0.860/0.884 & 0.867/0.886 & \textbf{0.861}/\textbf{0.888} & \textbf{0.891}/\textbf{0.915} & \textbf{0.923}/\textbf{0.936}\\
\midrule
 \multicolumn{10}{c}{\emph{AE: Kandinsky \citep{Razzhigaev23Kandinsky}}} \\ 
\midrule
$\text{AEROBLADE}_{\text{LPIPS}}$  & 0.801/0.756 & 0.879/0.891 & 0.976/0.977 & 0.864/0.870 & 0.686/0.715 & 0.691/0.718 & 0.643/0.600 & 0.715/0.735 & 0.782/0.783 \\
$\text{AEROBLADE}_{\text{LPIPS}_{2}}$ & 0.865/0.848 & 0.981/0.987 & 0.989/0.990 & 0.866/0.873 & 0.665/0.695 & 0.665/0.693 & 0.748/0.736 & 0.705/0.733 & 0.810/0.819\\
$\methodname_{\text{LPIPS}}$ (ours) & 0.899/0.880 & 0.992/0.994 & \textbf{0.996}/\textbf{0.997} & \textbf{0.876}/0.877 & \textbf{0.859}/\textbf{0.884} & \textbf{0.864}/\textbf{0.886} & 0.826/0.809 & \textbf{0.853}/\textbf{0.872} & \textbf{0.895}/\textbf{0.900} \\
$\methodname_{\text{LPIPS}_{2}}$ (ours) & \textbf{0.938}/\textbf{0.948} & \textbf{0.993}/\textbf{0.996} & 0.994/0.996 & 0.871/\textbf{0.883} & 0.768/0.797 & 0.773/0.796 & \textbf{0.895}/\textbf{0.922} & 0.806/0.834 & 0.880/0.897\\
\midrule
 \multicolumn{10}{c}{\emph{AE: MiniSD \citep{MiniSD}}} \\ 
\midrule
$\text{AEROBLADE}_{\text{LPIPS}}$  & 0.723/0.676 & 0.678/0.643 & 0.922/0.903 & 0.919/0.919 & 0.980/0.982 & 0.979/0.982 & 0.661/0.621 & 0.982/0.985 & 0.855/0.839\\
$\text{AEROBLADE}_{\text{LPIPS}_{2}}$ & 0.754/0.687 & 0.827/0.838 & 0.960/0.955 & \textbf{0.930}/\textbf{0.932} & 0.986/0.989 & 0.986/0.989 & 0.625/0.554 & 0.988/0.991 & 0.882/0.867\\
$\methodname_{\text{LPIPS}}$ (ours) & \textbf{0.834}/\textbf{0.823} & 0.866/0.885 & 0.979/0.976 & 0.890/0.868 & 0.997/0.998 & \textbf{0.998}/\textbf{0.998} & \textbf{0.853}/\textbf{0.858} & \textbf{0.998}/0.998 & 0.927/0.925\\
$\methodname_{\text{LPIPS}_{2}}$ (ours) & 0.778/0.735 & \textbf{0.951}/\textbf{0.966} & \textbf{0.987}/\textbf{0.982} & 0.921/0.918 & \textbf{0.998}/\textbf{0.999} & \textbf{0.998}/\textbf{0.998} & 0.808/0.822 & \textbf{0.998}/\textbf{0.999} & \textbf{0.930}/\textbf{0.927}\\
\bottomrule
\end{tabular}}
\end{table*}

\begin{table*}[t]
\centering
\caption{Cross-autoencoder AI-generated image detection performance (AUROC/AUPR) of \methodname and AEROBLADE \citep{ricker24aeroblade} in the DiffusionFace \citep{Chen24diffusionface} dataset. \textbf{Bold} denotes the best method.}
\label{table:crossae_diffusionface_appendix}
\resizebox{\textwidth}{!}{
\begin{tabular}{cccccc}
\toprule
                                 
Method  & SDv1.5 T2I & SDv2.1 T2I & SDv1.5 I2I & SDv2.1 I2I & Mean \\ 
\midrule 
\multicolumn{6}{c}{\emph{AE: SDv1.4 \citep{Rombach22SD}} } \\ 
\midrule
$\text{AEROBLADE}_{\text{LPIPS}}$  & 0.884/0.860 & 0.507/0.477 & 0.978/0.972 & \textbf{0.630}/\textbf{0.612} & 0.750/0.730 \\
$\text{AEROBLADE}_{\text{LPIPS}_{2}}$ & 0.865/0.833 & 0.464/0.452 & 0.986/0.982 & 0.600/0.586 & 0.729/0.713\\
$\methodname_{\text{LPIPS}}$ (ours) & \textbf{0.922}/\textbf{0.925} & \textbf{0.560}/\textbf{0.583} & 0.980/0.979 & 0.622/0.591 & \textbf{0.772}/\textbf{0.770}  \\
$\methodname_{\text{LPIPS}_{2}}$ (ours) & 0.899/0.873 & 0.513/0.523 & \textbf{0.993}/\textbf{0.992} & 0.605/0.582 & 0.753/0.743\\
\midrule
 \multicolumn{6}{c}{\emph{AE: SDv2-base \citep{Rombach22SD}}} \\ 
\midrule
$\text{AEROBLADE}_{\text{LPIPS}}$ & 0.852/0.831 & 0.514/0.486 & 0.837/0.821 & 0.631/\textbf{0.613} & 0.709/0.688  \\
$\text{AEROBLADE}_{\text{LPIPS}_{2}}$ & 0.848/0.813 & 0.487/0.467 & 0.875/0.861 & 0.609/0.595 & 0.705/0.684  \\
$\methodname_{\text{LPIPS}}$ (ours) & 0.860/\textbf{0.865} & \textbf{0.593}/\textbf{0.578} & 0.888/0.873 & \textbf{0.640}/0.600 & 0.732/\textbf{0.729} \\
$\methodname_{\text{LPIPS}_{2}}$ (ours) & \textbf{0.871}/0.843 & 0.518/0.532 & \textbf{0.933}/\textbf{0.925} & 0.622/0.595 & \textbf{0.736}/0.724\\
\midrule
\multicolumn{6}{c}{\emph{AE: Kandinsky \citep{Razzhigaev23Kandinsky}}} \\ 
\midrule
$\text{AEROBLADE}_{\text{LPIPS}}$  & 0.846/0.827 & 0.509/0.484 & 0.637/0.619 & 0.611/\textbf{0.595} & 0.651/0.631\\
$\text{AEROBLADE}_{\text{LPIPS}_{2}}$ & 0.845/0.812 & 0.466/0.453 & 0.692/0.666 & 0.589/0.578 & 0.648/0.627\\
$\methodname_{\text{LPIPS}}$ (ours) & \textbf{0.925}/\textbf{0.925} & \textbf{0.601}/\textbf{0.644} & 0.760/0.735 & \textbf{0.623}/0.586 & \textbf{0.727}/\textbf{0.723}\\
$\methodname_{\text{LPIPS}_{2}}$ (ours) & 0.882/0.858 & 0.505/0.514 & \textbf{0.779}/\textbf{0.758} & 0.594/0.575 & 0.690/0.676\\
\midrule
 \multicolumn{6}{c}{\emph{AE: MiniSD \citep{MiniSD}}} \\ 
\midrule
$\text{AEROBLADE}_{\text{LPIPS}}$  & 0.838/0.803 & 0.517/0.478 & 0.439/0.456 & 0.610/0.594 & 0.601/0.583 \\
$\text{AEROBLADE}_{\text{LPIPS}_{2}}$ & 0.844/0.808 & 0.506/0.467 & 0.489/0.492 & \textbf{0.615}/\textbf{0.601} & 0.614/0.592\\
$\methodname_{\text{LPIPS}}$ (ours) & \textbf{0.918}/\textbf{0.915} & \textbf{0.667}/\textbf{0.646} & 0.587/0.571 & 0.599/0.577 & \textbf{0.693}/0.677\\
$\methodname_{\text{LPIPS}_{2}}$ (ours) & 0.917/0.907 & 0.605/0.577 & \textbf{0.638}/\textbf{0.637} & 0.609/0.589 & 0.692/\textbf{0.678} \\
\bottomrule
\end{tabular}}
\end{table*}

\begin{table*}[t]
\centering
\caption{Ablation studies of \methodname with discrete cosine transform as $\text{F}$ under different hyperparameters on cutoff frequency $f$. We report the ensemble performance. \textbf{Bold} denotes the best hyperparameter choice.}
\label{table:appendix_dct_hyperparameter}
\resizebox{\textwidth}{!}{
\begin{tabular}{cccccccccc}
\toprule
                                 
$f$ & ADM & BigGAN & GLIDE & Midj & SD1.4 & SD1.5 & VQDM & Wukong & Mean \\ 
\midrule 
8 & \textbf{0.739} & \textbf{0.900} & \textbf{0.972} & \textbf{0.973} & 0.972 & 0.974 & 0.603 & 0.975 & \textbf{0.889}\\
16 & 0.723 & 0.886 & 0.965 & 0.968 & 0.972 & 0.973 & 0.600 & 0.977 & 0.883\\
24 & 0.727 & 0.898 & 0.968 & 0.965 & 0.972 & 0.973 & 0.613 & 0.979 & 0.887\\
32 & 0.719 & 0.896 & 0.969 & 0.962 & 0.974 & 0.975 & 0.621 & 0.982 & 0.887\\
36 & 0.709 & 0.888 & 0.969 & 0.960 & 0.976 & 0.977 & 0.632 & 0.984 & 0.887 \\
40 & 0.700 & 0.877 & 0.967 & 0.956 & 0.978 & 0.978 & 0.632 & 0.986 & 0.884\\
44 & 0.690 & 0.868 & 0.965 & 0.950 & 0.979 & 0.980 & 0.633 & 0.987 & 0.881\\
48 & 0.681 & 0.847 & 0.962 & 0.940 & \textbf{0.981} & \textbf{0.981} & \textbf{0.635} & \textbf{0.988} & 0.877\\
\bottomrule
\end{tabular}}
\vskip -0.1in
\end{table*}

We refer to Table \ref{table:crossae_genimage_appendix} and \ref{table:crossae_diffusionface_appendix} for the full results of \methodname compared to AEROBLADE. 

\section{Further details on the experiment}

\textbf{Datasets.} For the further information, diffusionface's T2I and I2I datasets constitutes of 90000 data each. For fair comparison against the real dataset with the 30000 dataset, we take a subset of the diffusionface with the 30000 dataset. We provide each link for the specified dataset we used in the footnote for SDv1.5 T2I \footnote{\url{stable_diffusion_v_1_5_text2img_p3g7.tar}}, SDv2.1 T2I\footnote{\url{stable_diffusion_v_2_1_text2img_p0g5.tar}}, SDv1.5 I2I \footnote{\url{SDv15_DS0.3.tar}}, and SDv2.1 I2I\footnote{\url{SDv21_DS0.3.tar}} dataset all downloadable through the author's official GitHub repository \footnote{\url{https://github.com/Rapisurazurite/DiffFace}}. The whole GenImage dataset is downloaded through the author's GitHub repository \footnote{\url{https://github.com/GenImage-Dataset/GenImage}}.

\textbf{Baselines.} First, we elaborate the specified link we used to evaluate the training-based AI-generated image detection methods in the footnote for DRCT \citep{Chen24DRCT} \footnote{\url{https://github.com/beibuwandeluori/DRCT}} and NPR \citep{Tan24NPR} \footnote{\url{https://github.com/chuangchuangtan/NPR-DeepfakeDetection}}. Note that we directly use the author's checkpoint following the data augmentation strategies noted in the corresponding paper. Furthermore, for evaluating the baseline performance of LatentTracer \citep{Wang24LatentTracer}, we use the author's official code \footnote{\url{https://github.com/ZhentingWang/LatentTracer}}. 

\section{Further hyperparameter search results on the discrete cosine transform}

We further explore the available choices of the hyperparameter $f$ which corresponds to the cutoff frequency of erasing high-frequency components in the frequency map computed by the discrete cosine transform. Note that $f=36$ corresponds to the Nyquist frequency $f=64/\sqrt{3}\approx 36.95$.

\end{document}